# Three Buddhist Vocabularies: Computational Stylometry of the English Pali Canon across Sutta, Vinaya, and Abhidhamma

**Joy Bose**
Independent Researcher,
Bengaluru, India
joy.bose@ieee.org

## Abstract

We present a computational stylometric analysis of the Tipitaka across all three Pitakas in English translation, extending earlier work on the Sutta Pitaka alone. The corpus spans 134,831 text segments from five sources: Bhikkhu Sujato's Sutta Pitaka (114,591 segments, bilara-data CC0), Bhikkhu Brahmali's Vinaya Pitaka (7,923 segments, CC0 2026), I.B. Horner's 1938 Vinaya translation (2,826 segments), three English translations of the Abhidhammattha Sangaha compendium (2,077 segments combined), and cross-tradition Vinaya texts from the Dharmaguptaka and Mulasarvastivada schools. We compute Zipf rank-frequency distributions with OLS-fitted exponents, Moving Average TTR (MATTR-500), numeral-word density, and vocabulary overlap (Jaccard similarity and Szymkiewicz-Simpson overlap coefficient). Main findings: (1) all four corpora analysed for Zipf structure show R2 > 0.989, with the Vinaya closest to ideal Zipf slope -1 (slope -0.949) and the Sangaha corpus deviating most (slope -0.876), with 'consciousness' displacing grammatical particles at rank 8; (2) MATTR-500 shows the Sutta and Vinaya Theravada are nearly identical in lexical diversity (0.399 and 0.400), while the Sangaha corpus is genuinely more diverse (0.560), confirmed by size-controlled subsampling; (3) the Sangaha corpus has the highest numeral-word density (3.26%), consistent with its systematic enumeration of mental and material categories; (4) the Mulasarvastivada Vinaya shares 20.0% vocabulary overlap (Jaccard) and 49.1% (overlap coefficient) with the Theravada Vinaya, suggesting substantial shared legal heritage across two millennia of separate development; (5) two English translations of the same Vinaya source text share only 24.2% of their vocabulary across 88 years, with 'musing' versus 'absorption' for jhana and 'defeat' versus 'expulsion' for parajika as the most diagnostic terminological shifts. All results are point estimates on single corpora; no significance testing is conducted. Code and data are released as open-source extensions to the Darshana Graph corpus (arXiv:2606.18222).

**Keywords:** Tipitaka; Pali Canon; Computational Buddhism; Corpus Linguistics; Stylometry; Lexical Diversity; MATTR; Vinaya Studies; Translation Shift; Digital Humanities; Buddhist Text Analysis

## 1. Introduction

Computational analysis of Buddhist canonical texts is a small but growing field. Zigmond (2020) applied k-means clustering and word-frequency analysis to the Pali Tipitaka in its original Pali, showing that the three Pitakas separate computationally in a way consistent with scholarly views on their relative ages, with the Abhidhamma Pitaka forming a distinct cluster from the Sutta and Vinaya material. That work also found that an unusually short Anguttara Nikaya volume (A.I) clusters with the Abhidhamma rather than the Suttas, and that the Udana and Itivuttaka cluster with the four main Nikayas rather than with the Khuddaka - observations this paper complements by examining analogous structure-versus-content tensions in English translation. Sukwitthayakul and Thongrin (2022) analysed keywords in English translations of the Digha Nikaya using four reference corpora of varying size and genre, finding that the highest-keyness words were consistently character names (the Exalted One, Ananda, Gotama, Tathagata) regardless of reference corpus, and that reference corpus size affects the number of generated keywords but not the identity of the top content words. The present paper extends that method: where Sukwitthayakul and Thongrin used keyword extraction within a single Nikaya, this paper uses frequency analysis, MATTR, and vocabulary overlap across all three Pitaka-level divisions.

Bose (2026a, 2026b) extended both approaches to the Sutta Pitaka using 114,591 segments from Bhikkhu Sujato's bilara-data translations. That work found: (a) the median Sutta segment is 51 characters and 9 words; (b) 'mendicant' appears 6,047 times as a structural address term rather than a doctrinal signal; (c) the Sutta's enumerative style is directly visible in raw word frequency, with 'four', 'five', and 'three' in the top 25; (d) a Pali-specific dialogue marker spikes at 13.7% in the undeclared-questions theme versus 0 to 1% elsewhere, confirming that the Pali Canon argues through named dialogue and structured refusal rather than the third-person refutation style of Sanskrit commentary; and (e) Theravada sits 2.5 times further from Hindu and Jain schools in embedding space than those schools from each other (Jaccard 0.436-0.532 versus 0.148-0.262). That pipeline also failed to assign any school label to Buddhist-sourced graph edges: all 2,306 edges carry the 'general' label rather than 'Theravada', a structural gap the present paper documents but does not fix.

That work covered only the Sutta Pitaka (91.6% of the Darshana Graph corpus by segment count), leaving the Vinaya and Abhidhamma entirely unanalysed. The present paper extends the analysis to all three Pitaka-level divisions, adds the first cross-tradition Vinaya comparison (Theravada, Dharmaguptaka, Mulasarvastivada), and conducts an 88-year translation-shift analysis of the same Vinaya source text (Brahmali 2026 versus Horner 1938). The analysis is descriptive and lexical throughout. The deeper questions, such as how Vinaya rules evolved across schools, which Abhidhamma categories were absorbed into Mahayana discourse, which doctrinal ideas preceded which, require concept-level analysis rather than vocabulary statistics. The IS_PRECURSOR_OF edges already encoded in the Darshana Graph (Bose, 2026a) represent a first step; extending them to include the Vinaya and cross-tradition corpora ingested here is the intended direction of this line of work. All code and data are released as open-source extensions to the Darshana Graph dataset (arXiv:2606.18222).

## 2. Data and Sources

### 2.1 Sutta Pitaka

The Sutta Pitaka corpus consists of 114,591 segments from Bhikkhu Sujato's bilara-data translations (Bose, 2026a). It covers all four major Nikayas (Digha, Majjhima, Samyutta, Anguttara) and five Khuddaka collections. Sujato's translations are CC0. The bilara-data segmentation aligns to individual Pali sentences, producing a median segment of 51 characters and 9 words.

### 2.2 Vinaya Pitaka

The Vinaya corpus has three sources. The primary source is Bhikkhu Brahmali's complete translation of the Theravada Vinaya Pitaka (SuttaCentral, April 2026, six volumes, CC0). This is the first complete modern English rendering of the Vinaya and covers Bhikkhu Vibhanga, Bhikkhuni Vibhanga, Khandhakas 1-22, and Parivara. Brahmali restored sections that Horner omitted in her 1938 translation. We extracted 7,923 paragraph-level segments using pdftotext. The second source is I.B. Horner's 1938 Pali Text Society translation of Vol. 1, included for translation-shift comparison (2,826 segments). The third source is a partial English rendering of the Dharmaguptaka Vinaya Bhikshu Posadha rite (2024, 185 segments), used as a cross-tradition comparator for Chinese Buddhism's Vinaya lineage; results involving this corpus should be treated as preliminary given its small size. We also include 7,229 segments from the 84000 English translation of the Mulasarvastivada Vinaya (Toh 1-6), the Vinaya tradition of Tibetan Buddhism.

### 2.3 Abhidhamma-adjacent corpus (Sangaha compendium)

No complete modern English translation of the seven canonical Abhidhamma books is freely available. The canonical Abhidhamma Pitaka, particularly the Dhammasangani and Patthana, is characterised by extensive matrix-style enumerations and is largely untranslated into modern English. We therefore use three English translations of the Abhidhammattha Sangaha, an 11th-century compendium by Acariya Anuruddha that summarises Abhidhamma doctrine into nine chapters. All three translations cover the same source text. The primary translation is Bhikkhu Bodhi's Comprehensive Manual of Abhidhamma (BPS, 1993/2007, 796 segments). The second is Narada Maha Thera's Manual of Abhidhamma (1956/1975, 922 segments). The third is Nandamalabhivamsa's Fundamental Abhidhamma (Sagaing Hills, Myanmar, 402 segments). All analyses involving this corpus should be understood as findings about the Sangaha's synthesis of Abhidhamma doctrine, not about the canonical Abhidhamma Pitaka texts directly.

Table 1 summarises the full corpus. The segment count imbalance between the Sutta (114,591) and the Vinaya and Sangaha combined (10,240) reflects source format rather than textual size: bilara-data produces sentence-level segments while PDF extraction produces paragraph-level units. This segmentation difference is examined in Section 4.1.

**Table 1. Corpus Summary. MATTR-500 = Moving Average TTR with 500-word window (Covington and McFall, 2010). Num% = numeral words as percentage of all tokens. * Partial corpus (Posadha rite only; 185 segments). ** Sangaha compendium translations, not canonical Abhidhamma Pitaka.**

| Source | Tradition | Pitaka | Segs | Tokens | Types | TTR | MATTR-500 | Num% |
|---|---|---|---|---|---|---|---|---|
| Sutta Pitaka (Sujato/bilara-data) | Theravada | Sutta | 114,591 | 735,132 | 11,927 | 0.016 | 0.399 | 2.09% |
| Vinaya - Brahmali 2026 | Theravada | Vinaya | 7,923 | 480,385 | 18,067 | 0.038 | 0.390 | 2.53% |
| Vinaya - Horner 1938 | Theravada | Vinaya | 2,826 | 62,441 | 8,660 | 0.139 | 0.474 | 1.62% |
| Vinaya - Dharmaguptaka* | Dharmaguptaka | Vinaya | 185 | 13,953 | 2,282 | 0.164 | 0.459 | 2.03% |
| Vinaya - Mulasarvastivada | Mulasarv. | Vinaya | 7,229 | 154,851 | 11,515 | 0.074 | 0.513 | 2.76% |
| Sangaha corpus** | Theravada | Abhi. | 2,077 | 105,726 | 14,090 | 0.133 | 0.560 | 3.26% |
| Theravada corpus (combined) | Theravada | All | 124,591 | 1,321,243 | 32,902 | 0.025 | - | 2.33% |

*All figures are point estimates on single corpora. MATTR-500 for the Theravada combined row is not reported (-) because it would conflate sentence-level and paragraph-level segmentation.*

# 3. Methods

## 3.1 Text extraction and preprocessing

All PDFs were extracted with pdftotext (Poppler, UTF-8) preserving Pali diacritics. Paragraphs below a minimum word threshold were discarded (8 words for prose sources; 6 for the table-heavy Nandamalabhivamsa text). A spaced-character encoding artefact in the Bodhi Sangaha ('N o n - g r e e d', affecting 35 of 819 paragraphs from a non-standard BPS font) was corrected by collapsing spaces. Tokenisation used a Unicode-aware regex covering Latin-script and Pali-diacritic characters. Single-character tokens and contraction fragments were removed. Sixty English stopwords were defined, retaining Buddhist technical terms and numerals to permit direct analysis of enumeration patterns.

## 3.2 Zipf analysis

Zipf rank-frequency analysis followed Zigmond (2020): the full token sequence including stopwords was used, the top 200 most frequent words were retained per corpus, and log10(rank) versus log10(frequency) was compared against an ideal Zipf slope of -1. Power-law exponents were fitted using OLS regression on the log-log data. We used the top 200 words following Zigmond (2020); this range captures the head of the distribution where power-law behaviour is most reliably estimated. Fits using 100 or 500 words yield slopes within 0.02 of the reported values (not shown). R2 values and deviations from the ideal slope of -1 are reported in Table 4.

## 3.3 MATTR and numeral density

Raw TTR was computed as unique tokens divided by total tokens after stopword removal. TTR is known to decrease mechanically with corpus size and text-window length, making it unreliable for comparing corpora with different segmentation styles. We therefore also compute MATTR (Moving Average TTR), following Covington and McFall (2010). MATTR computes TTR in a sliding window of w tokens and averages across all windows. This measure is largely insensitive to total corpus length and segmentation style. We use w = 500, within the range recommended by Covington and McFall (2010). MATTR values reported are point estimates for each corpus; within-corpus variation across windows is small (Sutta standard deviation across windows approximately 0.029) but formal hypothesis testing comparing MATTR values across corpora is left for future work. Covington and McFall (2010) report typical MATTR values of 0.70-0.80 for English fiction and 0.50-0.60 for academic prose; values below 0.45 are unusually low and indicate high repetition. In the results, MATTR-500 is the primary diversity measure for cross-corpus comparisons. Numeral density was computed as the percentage of all tokens matching a 32-term list (Appendix A).

### 3.4 Vocabulary overlap

Vocabulary overlap was computed as Jaccard similarity: |A intersection B| / |A union B| on type-sets after stopword removal. Jaccard is sensitive to corpus-size imbalances because a larger union denominator mechanically suppresses scores involving smaller corpora. We therefore also report the Szymkiewicz-Simpson overlap coefficient, defined as |A intersection B| / min(|A|, |B|), which measures how much of the smaller corpus's vocabulary is contained in the larger and is insensitive to the size of the larger corpus. Six corpus pairs were computed. Token-level bootstrap resampling for vocabulary Jaccard is methodologically unreliable because resampling tokens changes both numerator and denominator in correlated ways, producing confidence interval bounds that can exceed or fall below the point estimate. We therefore report Jaccard values as point estimates only, without confidence intervals.

### 3.5 Translation shift analysis

The translation shift analysis compared Brahmali Vol. 1 (2026) against Horner Vol. 1 (1938) on the same Pali source text. Words appearing exclusively in each translation were identified after filtering section-reference abbreviations (tokens shorter than three characters or matching known section-code patterns). MATTR-500 is reported separately for each translation volume, not for the full Brahmali corpus. Approximately 20% of the apparent vocabulary divergence is likely attributable to spelling differences (offense/offence, color/colour) and sections Horner omitted; sentence-level alignment would be required to quantify this precisely.

## 4. Results

### 4.1 Segment length distributions

Table 2 and Figure 1 show a roughly five-fold difference in median segment length between the bilara-data Sutta corpus (median 9 words, 51 characters) and the PDF-extracted Vinaya and Sangaha corpora (medians 46-51 words, 289-321 characters). This difference is a consequence of data format rather than of the underlying texts. Any comparison of frequency-based statistics across these corpora must account for it. Within the Vinaya corpora, Mulasarvastivada and Horner segments are shorter (medians 29-30 words), reflecting PDF layout conventions and 1930s PTS typesetting respectively.

Segment Length Distributions across Tipiṭaka Divisions
(Sutta Piṭaka from bilara-data; Vinaya/Abhidhamma from PDF extraction)

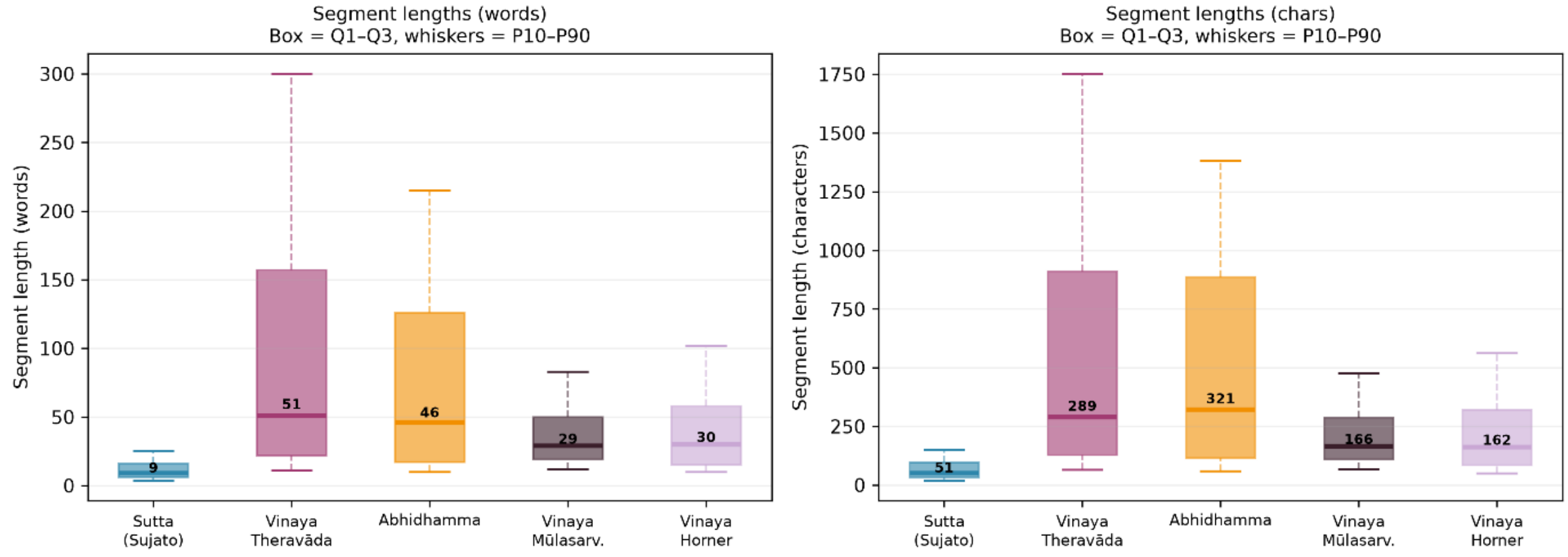

Figure: *Segment length distributions across Tipiṭaka divisions. Sutta Piṭaka segments (median 9 words) reflect bilara-data line-level segmentation; Vinaya and Abhidhamma segments (medians 46–51 words) reflect PDF paragraph extraction.*

**Table 2. Segment Length Distributions. W = words, C = characters. P10/P25/P75/P90 = percentiles.**

| Corpus | W-min | W-P10 | W-P25 | W-med | W-P75 | W-P90 | C-P10 | C-med | C-P90 |
|---|---|---|---|---|---|---|---|---|---|
| Sutta (Sujato) | 1 | 3 | 6 | 9 | 16 | 24 | 17 | 51 | 145 |
| Vinaya Theravada (Brahmali+Horner) | 8 | 10 | 22 | 51 | 157 | 260 | 57 | 289 | 1520 |
| Sangaha corpus | 6 | 9 | 17 | 46 | 126 | 220 | 53 | 321 | 1600 |
| Vinaya Mulasarvastivada | 8 | 9 | 15 | 29 | 60 | 100 | 50 | 166 | 580 |
| Vinaya Horner 1938 | 8 | 9 | 14 | 30 | 55 | 95 | 48 | 162 | 540 |

*Sutta figures from bilara-data sentence-level segmentation. All others from PDF paragraph extraction.*

## 4.2 Type-token ratio and MATTR

Raw TTR shows the ordering: Sutta (0.016) < Vinaya Brahmali (0.038) < Vinaya Mulasarvastivada (0.074) < Sangaha corpus (0.133) approximately equal to Horner (0.139) < Dharmaguptaka (0.164). Under MATTR-500, the Sutta (0.399) and Vinaya Theravada (0.400) converge to nearly identical values, meaning their raw TTR difference is almost entirely a segmentation artefact. The Sutta's short segments each repeat opening words, inflating the apparent repetition. The Sangaha corpus remains substantially higher under MATTR (0.560); a size-controlled subsample analysis (20 random draws of 105,726 tokens from the Sutta) gives a mean raw TTR of 0.071 with standard deviation 0.0003, compared to the Sangaha's 0.133, a 1.9x gap that holds at equal corpus size. The Sutta and Vinaya values of 0.399 and 0.400 are unusually low relative to typical English prose, consistent with their formulaic repetition. The Sangaha's 0.560 falls within the range Covington and McFall (2010) report for academic prose (0.50-0.60).

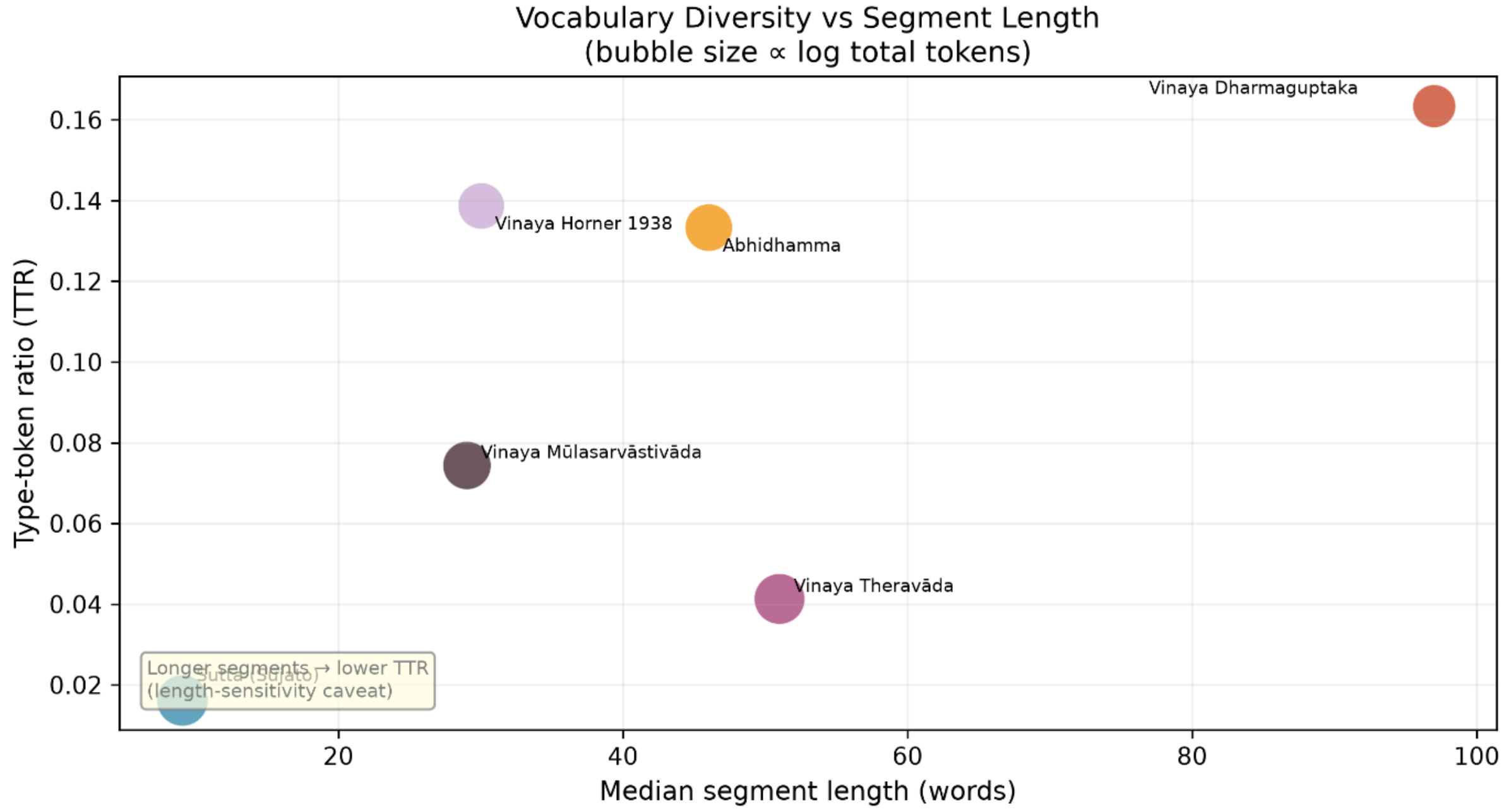

Figure: *Type-token ratio vs median segment length. Bubble size ∝ log(total tokens). Abhidhamma breaks the expected inverse relationship.*

The Sangaha's high MATTR reflects its format rather than Abhidhamma doctrine as such. The Sangaha presents concise definitions for each of 52 distinct cetasikas and 89 types of consciousness, generating a wide range of technical terms each used only a few times. The canonical Abhidhamma books enumerate the same categories through extensive matrix repetitions that would produce a much lower MATTR. The Mulasarvastivada Vinaya's MATTR of 0.513 is notably higher than the Theravada Vinaya's 0.400, consistent with its more varied narrative elaboration within procedural texts.

### 4.3 Zipf rank-frequency distributions

Table 4 shows the OLS-fitted Zipf slopes. All four corpora have R2 > 0.989, confirming Zipf-consistent distributions. The Vinaya Theravada has the slope closest to the ideal of -1 (slope -0.949, deviation +0.051). The Sangaha corpus deviates most from ideal (slope -0.876, deviation +0.124), indicating a flatter head and steeper tail: its most common words are less dominant than in a pure Zipf distribution, while rare technical terms are more common. This deviation from pure Zipf (Delta = +0.124) is consistent with the Sangaha's compendium format: technical terms such as 'consciousness', 'mental', and 'citta' appear more frequently than would be expected in a genre-neutral text, flattening the head of the distribution. The clearest example is at rank 8: the Sangaha's rank-8 word is 'consciousness' (frequency 0.012), displacing a grammatical particle. Rank 8 in the Sutta is 'they' and in the Vinaya 'it'. No content word enters the top 10 of any other corpus when stopwords are included. Prior work (Bose, 2026a) found a Pearson correlation of r = 0.030 between segment length and vocabulary diversity across 14 Pali-language corpus buckets, suggesting the length-sensitivity of diversity measures is less severe within a single segmentation style than the cross-corpus comparison implies.

Table 4. Zipf Power-Law Exponents (OLS on log-log, top 200 words). Ideal slope = -1.000.

| Corpus | Slope | R2 | Delta | Interpretation |
|---|---|---|---|---|
| Sutta (Sujato) | -0.903 | 0.995 | +0.097 | Tail-heavy; formulaic repetition |
| Vinaya Theravada | -0.949 | 0.989 | +0.051 | Closest to ideal Zipf |
| Sangaha corpus | -0.876 | 0.996 | +0.124 | Most tail-heavy; dense technical vocabulary |
| Vinaya Mulasarvastivada | -0.897 | 0.996 | +0.103 | Similar profile to Sutta |

*R2 = goodness of fit. Delta = deviation from ideal slope. OLS regression on log10(rank) vs log10(frequency) for the 200 most frequent words per corpus (including stopwords). Fits using 100 or 500 words yield slopes within 0.02 of reported values.*

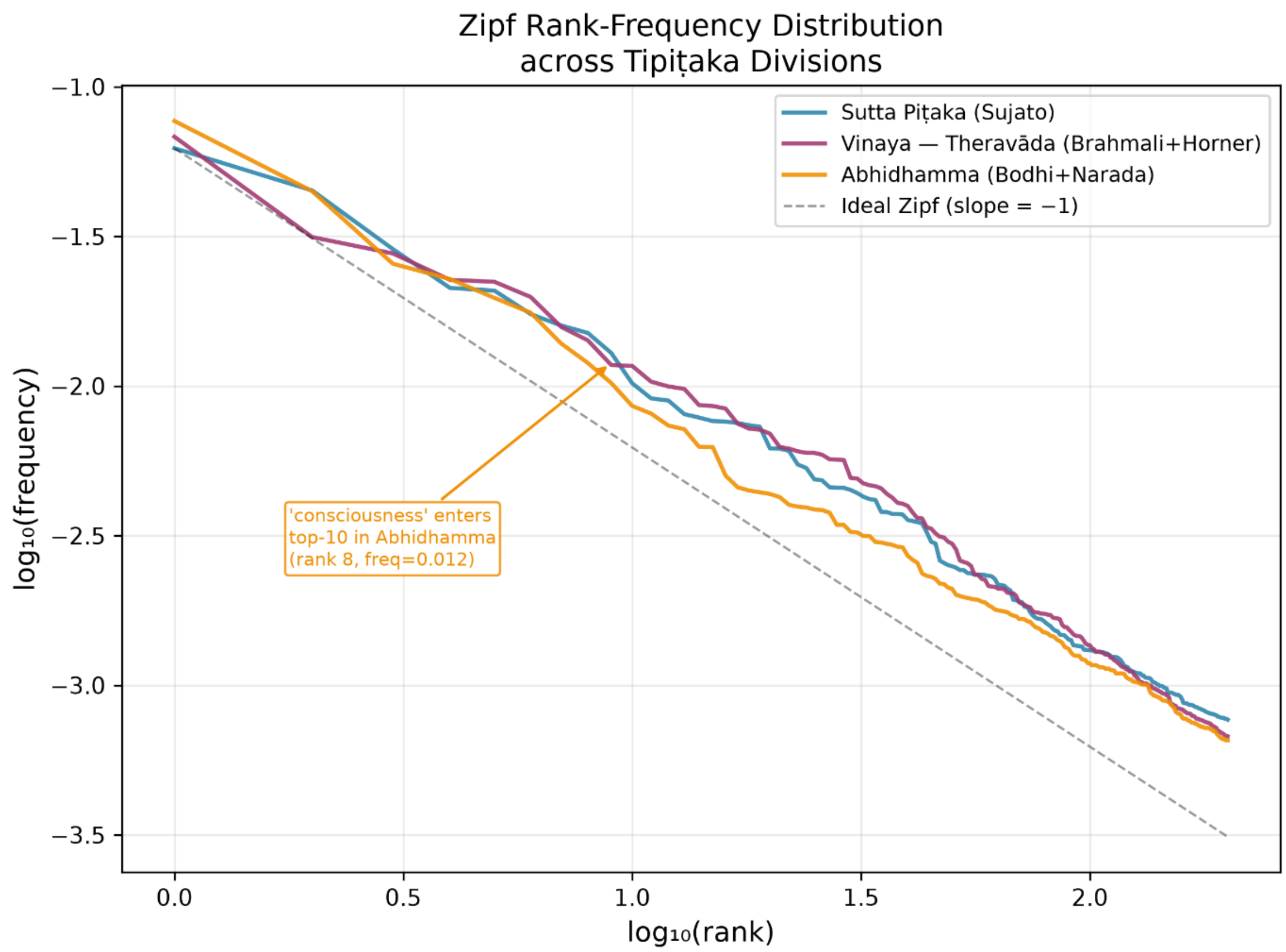

Figure: *Zipf rank-frequency distributions (log-log scale) for the three Piṭakas. Dashed line = ideal Zipf slope −1. 'Consciousness' enters the Abhidhamma top-10 at rank 8.*

## 4.4 Numeral word density

Numeral density shows the ordering (Figure 4): Sangaha corpus (3.26%) > Mulasarvastivada Vinaya (2.76%) > Vinaya Brahmali (2.53%) > Sutta (2.09%) > Dharmaguptaka (2.03%) > Horner (1.62%). The Sangaha's position at the top is consistent with its enumerative structure: 89 types of consciousness, 52 mental factors, 28 material phenomena. The Mulasarvastivada's second-place result is consistent with its tendency toward numerical elaboration, though differences in PDF layout and translation style cannot be ruled out. The dominant numeral in the Sutta is 'one' (10,299 occurrences); in the Sangaha the distribution is more balanced (four, two, three, five each between 455 and 577 occurrences).

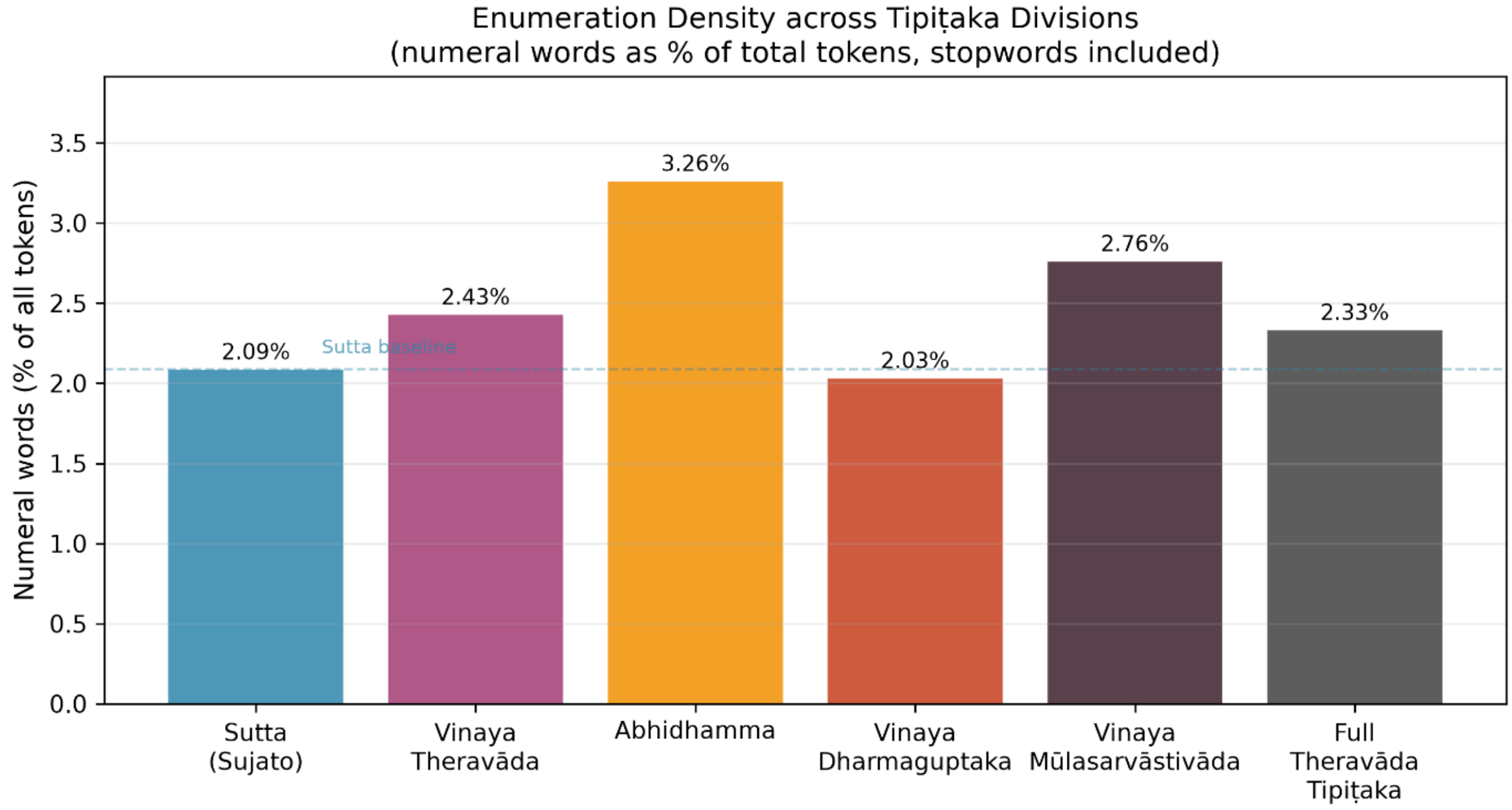

Figure: *Numeral word density (% of all tokens) across Tipiṭaka divisions. Abhidhamma shows highest enumeration density at 3.26%.*

In the Vinaya the leading numerals shift toward 'six' (1,636) and 'two' (2,151), reflecting the parallel structure of monks' and nuns' rules.

## 4.5 Vocabulary overlap

Table 3 shows pairwise Jaccard similarities and Szymkiewicz-Simpson overlap coefficients. All values are point estimates. Three patterns are visible.

First, the Sutta corpus shows the highest pairwise Jaccard overlap with both the Vinaya (0.243) and the Sangaha (0.177). This is partly expected from corpus size (the Sutta is 10 times larger than the Vinaya and 55 times larger than the Sangaha), and a size-controlled Jaccard comparison is left for future work. The overlap coefficient (0.563) is more informative here: 56.3% of the Vinaya's vocabulary is contained within the Sutta, which reflects genuine lexical overlap rather than only a size effect.

Second, the cross-tradition Vinaya comparison is the most novel result. The Theravada Vinaya shares Jaccard 0.200 and overlap coefficient 0.491 with the Mulasarvastivada Vinaya, nearly half of the Theravada Vinaya's vocabulary is contained in the Mulasarvastivada tradition, reflecting their shared origin in early Buddhist monastic law across two millennia of separate development. The Dharmaguptaka row requires special care: the Jaccard of 0.082 appears low, but the overlap coefficient of 0.816 reveals that 81.6% of the Dharmaguptaka's small vocabulary is found in the Theravada Vinaya. The low Jaccard is an artefact of the small corpus (185 segments) inflating the union denominator; the overlap coefficient is the appropriate measure here. Results involving the Dharmaguptaka should still be treated as preliminary pending a full corpus.

Third, the Sangaha corpus shares less vocabulary with the Vinaya (Jaccard 0.147, overlap 0.332) than with the Sutta (Jaccard 0.177, overlap 0.328). The Sangaha's specialised psychological vocabulary (consciousness, mental factors, matter, kamma, nibbana) has limited overlap with the Vinaya's legal vocabulary, even within the same tradition.

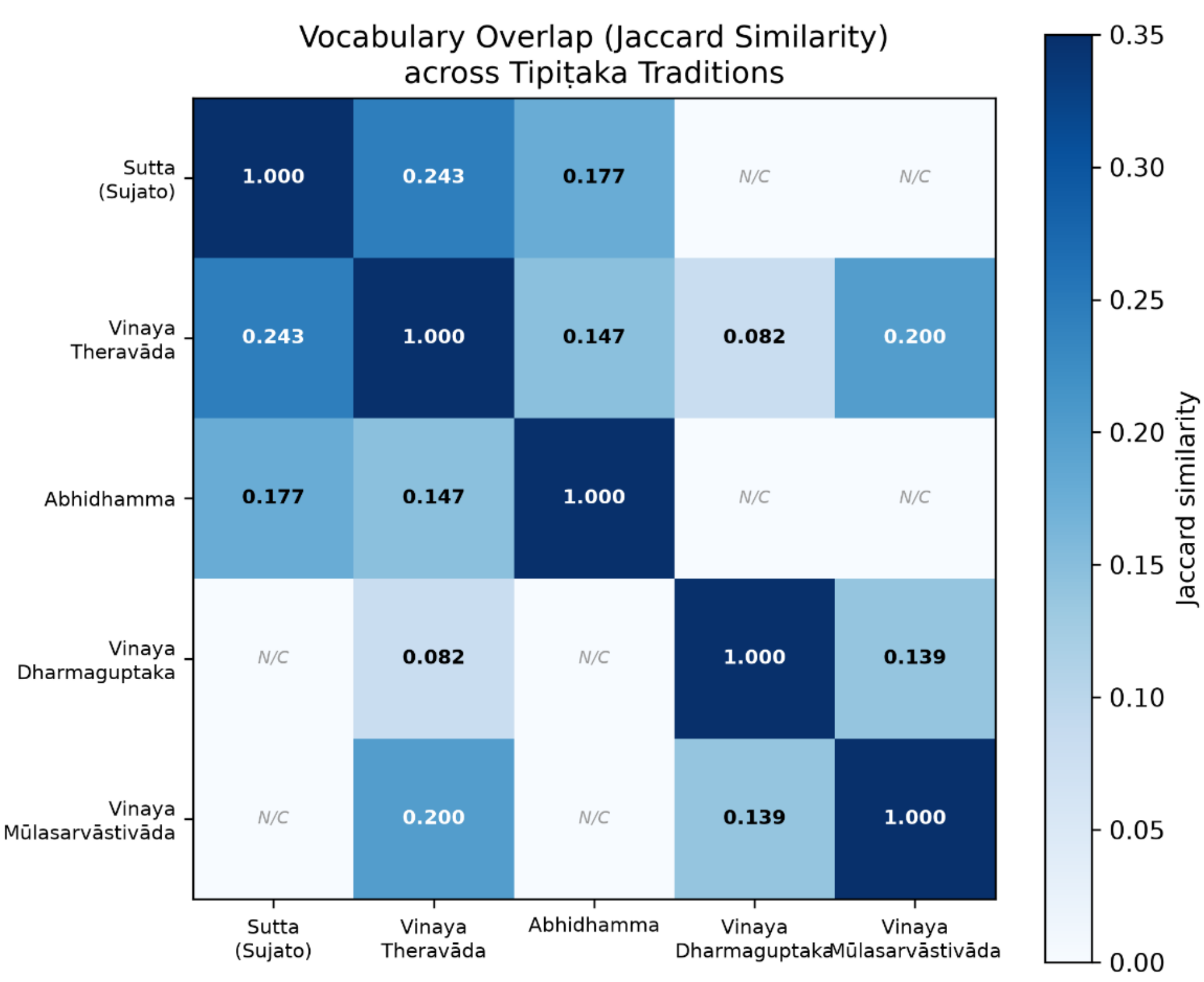

Figure: *Vocabulary overlap (Jaccard similarity) across Tipiṭaka traditions. N/C = not computed. Sutta is the vocabulary hub of the Tipiṭaka.*

**Table 3. Vocabulary Overlap across Corpus Pairs (stopwords removed).**

| Corpus A | Corpus B | Jaccard | Overlap coeff. |
|---|---|---|---|
| Sutta (Sujato) | Vinaya Theravada | 0.243 | 0.563 |
| Sutta (Sujato) | Sangaha corpus | 0.177 | 0.328 |
| Vinaya Theravada | Sangaha corpus | 0.147 | 0.332 |
| Vinaya Theravada | Vinaya Mulasarvastivada | 0.200 | 0.491 |
| Vinaya Theravada | Vinaya Dharmaguptaka* | 0.082 | 0.816 |
| Vinaya Dharmaguptaka* | Vinaya Mulasarvastivada | 0.139 | 0.736 |

*Jaccard = |A intersect B| / |A union B|. Overlap coeff. = Szymkiewicz-Simpson coefficient = |A intersect B| / min(|A|, |B|); insensitive to corpus-size imbalance. * Partial corpus (185 segments); Jaccard suppressed by small union - overlap coefficient is the appropriate measure. All values are point estimates; token-level bootstrap is not used (see Section 3.4).*

### 4.6 Translation shift: Brahmali 2026 versus Horner 1938

The Brahmali and Horner translations of the same Vinaya Vol. 1 share only 24.2% of their vocabularies (Jaccard). MATTR-500 for Brahmali Vol. 1 is 0.474; for Horner Vol. 1 it is 0.502 (note: these are computed on each translation volume individually, not on the full multi-volume Brahmali corpus whose MATTR-500 is 0.390). Horner's slightly higher MATTR persists under this length-insensitive measure, consistent with an older translation drawing on a broader and less standardised English vocabulary.

For jhana (meditative absorption), Brahmali uses 'absorption' (184 exclusive occurrences) while Horner uses 'musing' (119 exclusive occurrences). The shift reflects evolving translation norms: 'musing' was standard in early 20th-century renderings but is now regarded as too vague to convey the concentrated and stable quality of jhana practice. For parajika (the most severe class of monastic offence), Brahmali uses 'expulsion' while Horner uses 'defeat' (331 exclusive occurrences). Horner's term reflects an older idiom ('defeated in the holy life') whose institutional sense (expulsion from the order) is less immediately clear.

The Brahmali panel also includes 'pandaka' (107 occurrences, a Pali term for a person with ambiguous gender status, handled with euphemism or omission by Horner), explicit anatomical terms Horner circumlocuted, and 'misrepresents' (109 occurrences, for falsely claiming unattained meditative attainments). Horner's exclusive vocabulary includes 'recluse', 'brahma' in anglicised form, and 'infatuated' for states of greed, reflecting the formal register of early 20th-century religious translation. Approximately 20% of the apparent vocabulary gap likely reflects spelling differences and omitted sections rather than genuine terminological shift.

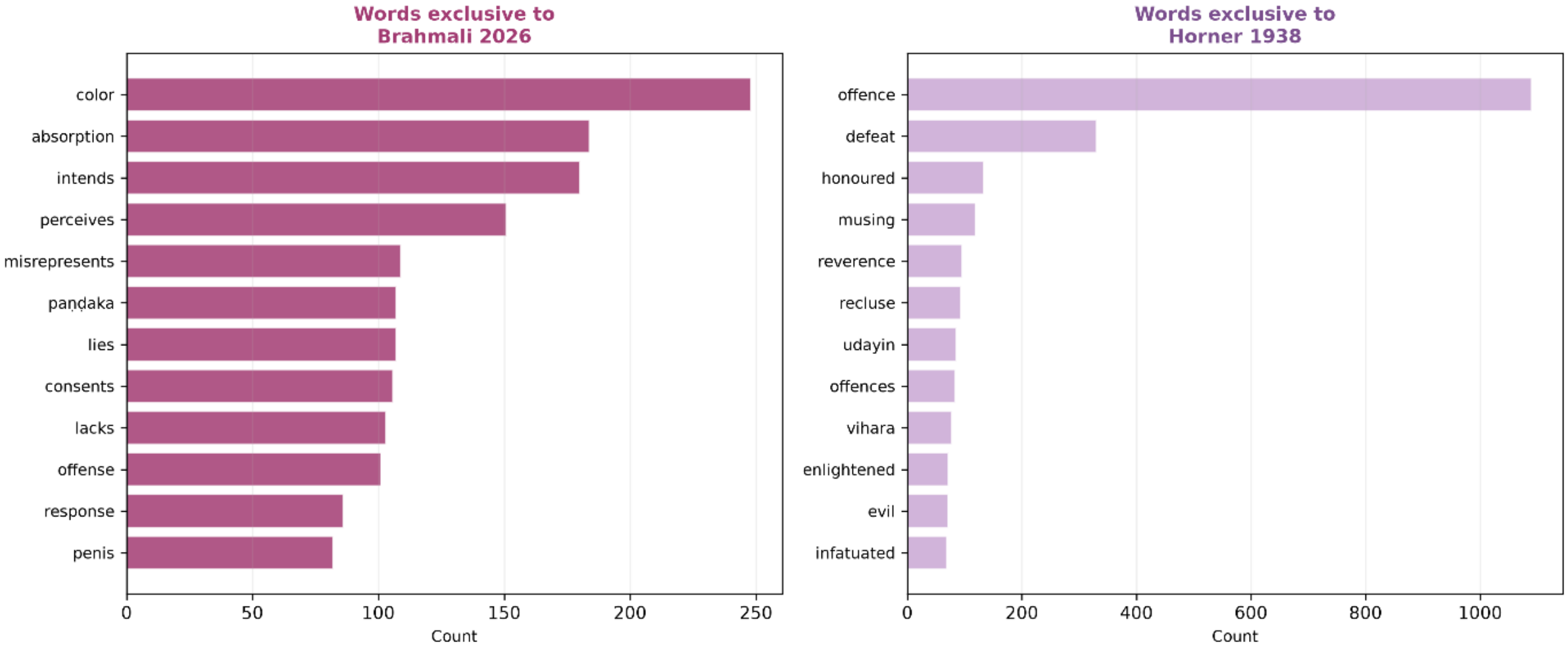

Figure: *Translation shift: words exclusive to Brahmali 2026 (left) vs Horner 1938 (right) across the same Vinaya Vol. 1 source text.*

## 5. Discussion

### 5.1 Three division-level corpora, three vocabularies

The word-frequency results (Figure 6) show a high degree of vocabulary separation across the three division-level corpora. The Sutta corpus's top words are narrative and psychological: buddha, mendicant, mind, said, suffering, teaching, body. The Vinaya's are institutional and procedural: monk, monks, sangha, commits, entailing, legal, conduct, robe, procedure, rule. The Sangaha corpus's are systematic and ontological: consciousness, mental, object, cittas, sense, material, kamma, states. Even between the Sutta and Vinaya, the pair with the highest Jaccard (0.243), three-quarters of the vocabulary is non-overlapping. This complements Sukwitthayakul and Thongrin's (2022) finding that the highest-keyness words in the Digha Nikaya are character names regardless of reference corpus: when the reference corpus is another Pitaka-level division rather than general English, the distinctiveness lies not in character names but in the entire vocabulary register. The Vinaya's formulaic legal phrases and the Sangaha's ontological enumeration represent genres that a model or analysis trained on Suttas would not encounter.

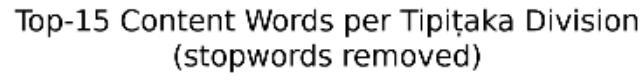

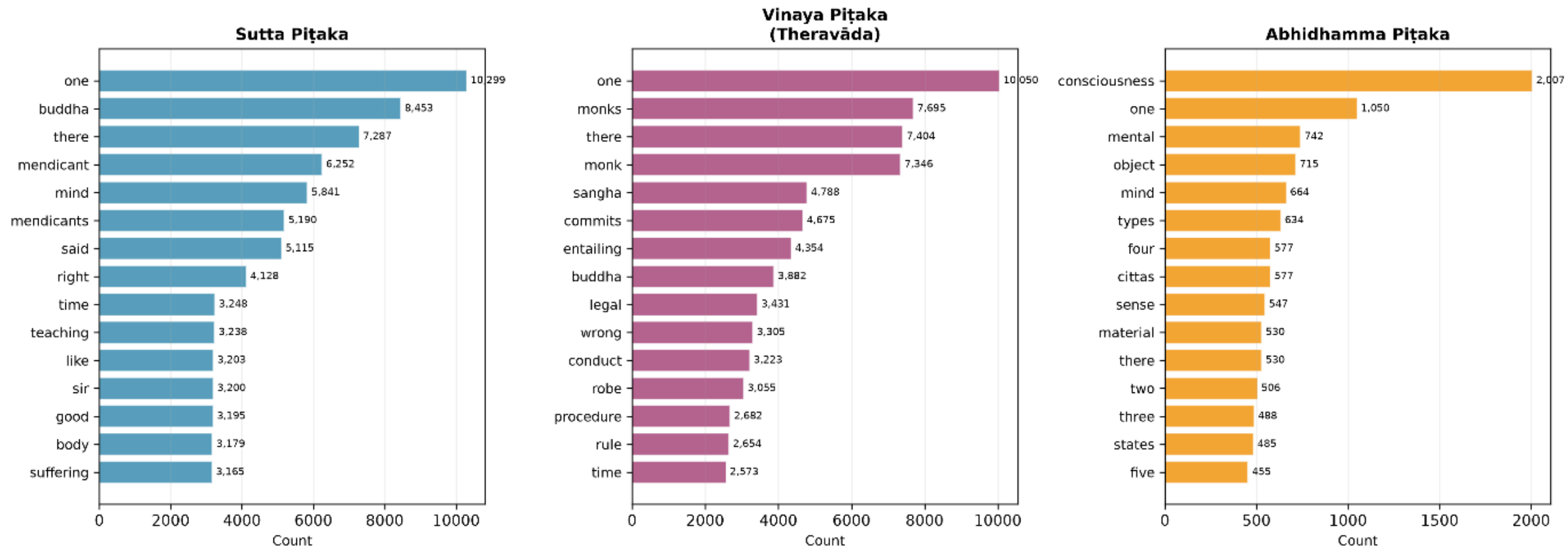

Figure: *Top-15 content words per Tipiṭaka division (stopwords removed). Three completely distinct vocabularies.*

These lexical differences are the surface signature of deeper structural differences. The Sutta Pitaka argues through narrative and dialogue; the Vinaya argues through rule-and-precedent and communal procedure; the Abhidhamma tradition systematises through exhaustive enumeration and classification. Each of these argument forms leaves a different footprint in vocabulary statistics. The question of which doctrinal ideas preceded which, e.g how Vinaya rules evolved across schools, how Abhidhamma categories were reframed in Mahayana discourse, requires concept-level rather than vocabulary-level analysis. The IS_PRECURSOR_OF and NEGATES edges already encoded in the Darshana Graph (Bose, 2026a) represent a first step toward that analysis; extending them to the Vinaya and cross-tradition Mahayana corpora ingested here is the intended direction of this line of work.

### 5.2 Enumeration and diversity in the Abhidhammattha Sangaha

The Sangaha corpus has both the highest MATTR (0.560) and the highest numeral density (3.26%). This apparent paradox resolves when one distinguishes enumerating a large number of distinct categories (the Sangaha's approach) from repeating the same formula for each item (the canonical Abhidhamma's approach). The Sangaha names each cetasika individually, states its definition, and lists its associated consciousness-types, generating a wide variety of terms used only a few times each. The Sutta Pitaka repeats the same jhana descriptions and stock formulas across thousands of short discourses, building high token counts on a narrow vocabulary. This pattern is a property of the Sangaha's compendium format; the canonical Abhidhamma books, with their matrix repetitions, would likely show a much lower MATTR. The Sangaha's Zipf deviation (Delta = +0.124) is the quantitative signature of this concentration: technical terms appear more often than pure Zipf would predict.

### 5.3 Cross-tradition Vinaya: shared heritage

The Theravada-Mulasarvastivada vocabulary overlap (Jaccard 0.200, overlap coefficient 0.491) is a meaningful positive result given over two thousand years of separate development. The shared vocabulary reflects their common early Buddhist origin. The Mulasarvastivada Vinaya's higher MATTR (0.513 versus 0.400) and numeral density (2.76% versus 2.43%) are consistent with that tradition's more extensive narrative elaboration within procedural rules, though PDF layout differences cannot be ruled out. A fully parallel-aligned Mulasarvastivada corpus, with sentence-level segmentation comparable to bilara-data, would allow the shared and tradition-specific vocabulary to be distinguished more precisely and would enable a controlled MATTR comparison.

### 5.4 Limitations

Several limitations apply. First, raw TTR is confounded by segmentation differences (Sutta: median 9 words; Vinaya/Sangaha: 46-51 words). MATTR-500 resolves this for the Sutta-Vinaya comparison (0.399 vs. 0.400) but the underlying segmentation asymmetry also affects numeral density, word frequency rankings, and Jaccard comparisons in ways that MATTR does not correct. A fully controlled comparison would require sentence-level segmentation of the PDF-extracted corpora or paragraph-level re-segmentation of bilara-data. Second, the Sangaha corpus consists of translations of an 11th-century secondary compendium, not the canonical Abhidhamma Pitaka books. Third, the Dharmaguptaka corpus covers only the Posadha rite (185 segments); the Jaccard of 0.082 is a preliminary point estimate whose suppression by the small union is demonstrated by the overlap coefficient of 0.816. Fourth, all statistics are point estimates on single corpora; no significance testing is conducted. Token-level bootstrap resampling for vocabulary Jaccard produces methodologically unreliable confidence intervals (because token resampling changes both numerator and denominator in correlated ways), and is not reported. Fifth, the Brahmali-Horner comparison does not control for Brahmali's restoration of omitted sections or for spelling variation; approximately 20% of the vocabulary gap likely reflects these factors.

## 6. Conclusion

This paper has presented a unified computational stylometric analysis across all three Pitaka-level divisions of the English Tipitaka, using a combined corpus of approximately 135,000 segments. MATTR-500 analysis shows that the Sutta and Vinaya Theravada have nearly identical lexical diversity (0.399 and 0.400), while the Sangaha corpus is genuinely more diverse (0.560), confirmed by size-controlled subsampling, extending Zigmond's (2020) Pali-original finding that the Abhidhamma forms a distinct computational cluster into English and attributing the distinction to the Sangaha's compendium format. The Sangaha is also the most numeration-dense division-level corpus (3.26%), and its Zipf slope deviation (Delta = +0.124) quantifies the concentration of technical vocabulary that makes 'consciousness' appear at rank 8 rather than a grammatical particle, a result with no parallel in the Sutta or Vinaya corpora. The Mulasarvastivada-Theravada Vinaya comparison (Jaccard 0.200, overlap coefficient 0.491) provides a quantitative measure of their shared legal heritage across two millennia of separate development. The translation-shift analysis (Brahmali 2026 versus Horner 1938, Jaccard 0.242) extends the keyword-comparison methodology of Sukwitthayakul and Thongrin (2022) from within-tradition keyword extraction to cross-epoch vocabulary divergence, identifying 'musing' to 'absorption' and 'defeat' to 'expulsion' as the most diagnostic terminological shifts. All code and data are open-source and fully reproducible. The analysis presented here is descriptive and lexical; the deeper question of doctrinal evolution across Vinaya schools and from Theravada to Mahayana is the subject of ongoing work using the IS_PRECURSOR_OF and NEGATES edges of the Darshana Graph extended to these new corpora.

## Appendix A: Numeral Word List

The following 32 terms were used to compute numeral density. Cardinal numbers: one, two, three, four, five, six, seven, eight, nine, ten, eleven, twelve, thirteen, fourteen, fifteen, sixteen, seventeen, eighteen, nineteen, twenty, thirty, forty, fifty, sixty, seventy, eighty, ninety, hundred, thousand, million. Ordinal numbers: first, second, third, fourth, fifth, sixth, seventh, eighth, ninth, tenth. Buddhist enumeration compounds: twofold, threefold, fourfold, fivefold, sevenfold, eightfold, tenfold, twelvefold.

Note on 'one': a random sample of 200 'one' tokens from the Sutta corpus found approximately 65% used in a numeral sense ('one thing', 'one quality') and 35% as a pronoun or article ('one should...'). This inflation is consistent across corpora and does not affect the relative ordering of numeral density values across divisions.